\documentclass[runningheads]{llncs}
\usepackage[T1]{fontenc}
\usepackage{multirow, makecell}%
\usepackage{amsmath,amssymb,amsfonts}%
\usepackage{graphicx}
\usepackage{algorithm}%
\usepackage{algorithmicx}%
\usepackage{algpseudocode}%
\usepackage{listings}%
\usepackage{float}
\usepackage{tablefootnote}
\usepackage{subfigure}
\usepackage{tikz}
\usepackage{booktabs}\let\cline\cmidrule
\usepackage{xcolor}%
\usepackage{textcomp}%
\usepackage{manyfoot}%
\usepackage{capt-of,etoolbox}
\usepackage{cite}
\usepackage{array}
\usepackage{hyperref}
\hypersetup{
    colorlinks=true,
    citecolor=blue,
    linkcolor=blue,
    filecolor=magenta,      
    urlcolor=cyan
}
\def\checkmark{\tikz\fill[scale=0.3](0,.35) -- (.25,0) -- (1,.7) -- (.25,.15) -- cycle;}
\begin{document}
\title{CrackUDA: Incremental Unsupervised Domain Adaptation for Improved Crack Segmentation in Civil Structures}
\titlerunning{CrackUDA}
%
\author{Kushagra Srivastava\and
Damodar Datta Kancharla \and
Rizvi Tahereen \and Pradeep Kumar Ramancharla \and Ravi Kiran Sarvadevabhatla \and Harikumar Kandath}
\authorrunning{K. Srivastava et al.}
%
\institute{International Institute of Information Technology, Hyderabad, India}
\maketitle              
\begin{abstract}
Crack segmentation plays a crucial role in ensuring the structural integrity and seismic safety of civil structures. However, existing crack segmentation algorithms encounter challenges in maintaining accuracy with domain shifts across datasets. To address this issue, we propose a novel deep network that employs incremental training with unsupervised domain adaptation (UDA) using adversarial learning, without a significant drop in accuracy in the source domain. Our approach leverages an encoder-decoder architecture, consisting of both domain-invariant and domain-specific parameters. The encoder learns shared crack features across all domains, ensuring robustness to domain variations. Simultaneously, the decoder's domain-specific parameters capture domain-specific features unique to each domain. By combining these components, our model achieves improved crack segmentation performance. Furthermore, we introduce BuildCrack, a new crack dataset comparable to sub-datasets of the well-established CrackSeg9K dataset in terms of image count and crack percentage. We evaluate our proposed approach against state-of-the-art UDA methods using different sub-datasets of CrackSeg9K and our custom dataset. Our experimental results demonstrate a significant improvement in crack segmentation accuracy and generalization across target domains compared to other UDA methods - specifically, an improvement of 0.65 and 2.7 mIoU on source and target domains respectively. Additional details and code can be accessed from \href{https://crackuda.github.io/}{https://crackuda.github.io}

\end{abstract}
\section{Introduction}\label{sec1}

\begin{figure*}[!ht]
\centering
\includegraphics[width=0.9\textwidth, height=4cm]{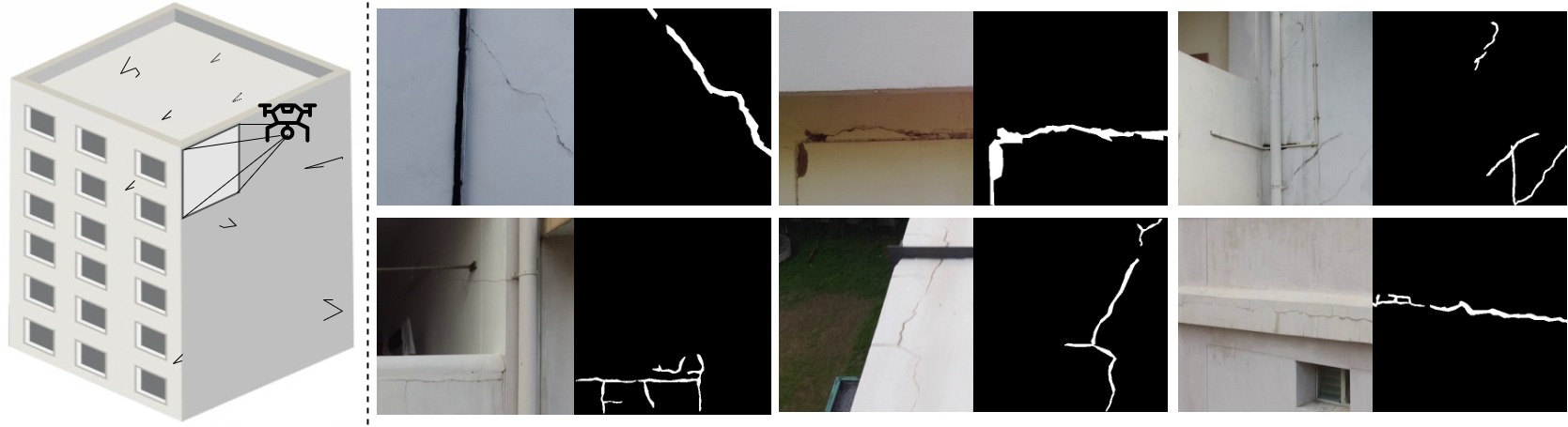}
 \caption{BuildCrack dataset was captured by imaging building facades using a drone-mounted camera from different angles and distances. BuildCrack has images with low contrast, occlusions, and shadows, which challenge the model's robustness. Sample images from our building crack dataset are shown. This dataset will be made public.}
        \label{fig:iiitdataset}
\end{figure*}

Identifying cracks in structures such as roads, pavements, and buildings is an important civil engineering task. This is especially crucial in determining a building's structural health and risk of failure during seismic activity~\cite{primer}. This task is being increasingly performed using visual imagery. However, the small footprint of cracks relative to building size and lack of regular structure make crack localization a challenging image segmentation problem. Different approaches have been explored over the years, ranging from rule-based to data-driven methods, for crack segmentation. Data-driven methods~\cite{lateef2019survey_segmentation} have gained prominence with the rise in the availability of crack datasets~\cite{rissbilder,crack500,volker2015crack,deepcrack,crackseg9k}. These methods have shown remarkable results in segmentation tasks. 

However, a key limitation of these approaches is their poor generalization across different domains, as datasets from various sources often have different distributions. This lack of generalization is evident when a model trained on one domain (source domain) is applied to a dataset from a different domain (target domain).
Several factors contribute to the domain shift observed in crack datasets. These include differences in image features, such as the contrast between cracks and their background, variations in crack shapes due to surface textures and lighting conditions, and the overall appearance of cracks~\cite{oliveira2017road,lau2020automated}. 

To address this challenge, domain adaptation techniques can be employed to reduce the domain shift. It is a viable solution since it alleviates the need for costly and labor-intensive annotation of crack segmentation data. Unsupervised Domain Adaptation (UDA) is a specific approach that adapts a network trained on a labeled source dataset to an unlabeled target dataset, effectively mitigating the problems associated with domain shift across datasets and high annotation costs ~\cite{intrada,fada,adaptsegnet,crst,advent,daformer,maxsquare,dacs}. While these UDA approaches have been extensively tested in domain adaptation tasks using real and synthetic autonomous driving datasets, our work demonstrates that these methods do not yield satisfactory results for the challenging crack segmentation setting

Our approach is designed to address the challenges of crack segmentation through an incremental learning setting. We employ a two-step process to adapt our network, trained on a labeled source dataset, to an unlabeled target dataset. To overcome catastrophic forgetting often observed in incremental learning approaches~\cite{forgetting}, our network architecture learns both domain-invariant and domain-specific feature representations. Our paper makes the following key contributions:

\begin{itemize}
    \item We propose CrackUDA, a novel incremental UDA approach that ensures robust adaptation and effective crack segmentation (Section \ref{sec:methodology}).   
    \item We demonstrate the effectiveness of CrackUDA by achieving higher accuracy in the task of building crack segmentation, surpassing the state-of-the-art UDA methods. Specifically, CrackUDA yields an improvement of 0.65 and 2.7 mIoU on the source and target domains, respectively (Section \ref{sec:results}).
    \item We introduce BuildCrack, a new building crack dataset collected via a drone (Section \ref{sec:results}).
\end{itemize}

\section{Related Works}

Crack segmentation approaches can be broadly categorized into two types: (i) rule-based and (ii) data-driven methods. Rule-based methods use human-defined rules to make decisions. Most of the rule-based methods have low accuracy because of non-uniform backgrounds, varying light conditions, and the brittle nature of the parameters \cite{KOCH2015196}. Data-driven methods leverage data samples to learn patterns and adjust the parameters of a model for a specific task. In particular, deep learning-based methods have demonstrated significant potential in crack segmentation and can be divided into supervised, weakly supervised, and semi-supervised based on the extent of supervision. 

Crack segmentation is dominated by supervised learning approaches. Encoder-decoder architecture \cite{deepcrack,deeplab,cracktree} has been popular for excellent performance in pixel-wise segmentation, provided accurately labeled segmentation maps are available. The encoder downsamples the input images to form a high-dimensional feature vector while the decoder reconstructs unique segmentation maps using this feature vector. CrackNet \cite{ zhang2017automated_cracknet} modifies the encoder-decoder architecture by using same-size convolution filters across layers to maintain explicit pixel-pixel representation. DeepCrack\cite{deepcrack} uses a fully convolution network (FCN) architecture with additional convolution layers at the end of a traditional CNN which upsamples feature maps of different scales to the original size and recovers fine-grained structures. CrackSeg9K \cite{crackseg9k} compared different state-of-the-art segmentation models. It was concluded that DeepLab v3+ \cite{deeplab} with ResNet and XceptionNet as the backbones worked best on linear cracks but the accuracy drops on webbed and branched cracks. With the introduction of vision transformer \cite{vit}, self-attention has become an efficient tool for learning non-local features. Crackformer \cite{liu2021crackformer} uses sequential self-attention networks for crack segmentation. The performance of supervised segmentation approaches relies on accurate semantic labels. Such approaches seldom generalize to datasets of different domains. \cite{curvilinear} proposed a curvilinear structure segmentation approach for crack segmentation on diverse datasets such as Crack500\cite{crack500} and CrackTree200\cite{cracktree}.

 \cite{konig2022weakly_weaksup_seg} propose a weakly supervised approach that uses inferior quality labels for crack segmentation. They demonstrated their network's capability to perform in out-of-domain (OOD) cases, but accuracy suffers when there are thin cracks in the target dataset. Semi-supervised approaches have used generative adversarial networks \cite{semi_crackseg} and super-resolution \cite{kondo2021crack_cssr} to generate pseudo-labels for training their network. However, the performance of these methods depends on the quality of the pseudo-labels. Though semi-supervised approaches perform well in the case of OOD, they require some labeled data of the target domain.

Since the conspicuous hurdle is reliable labeled data and poor generalization, our work uses UDA. UDA has demonstrated its potential for various vision tasks such as object detection~\cite{od2, od3, od4, od5}, classification~\cite{classification1, classification2,classification3, classfication4,classification5}, and more relevantly, semantic segmentation \cite{intrada,fada,adaptsegnet,crst, cbst,daformer,maxsquare,dacs} as well as crack segmentation \cite{weng2023unsupervised}.

\section{Preliminaries}

In this section, we formally define our problem statement and provide an overview of UDA and incremental learning.

\subsection{Problem Statement}

Consider a source distribution $S$ and target distribution $T$, both defined on the input-label space $X \times Y$. In this setting, $X \in \mathbb{R}^{H \times W \times 3}$ represents RGB images, while $Y \in \mathbb{R}^{H \times W}$ corresponds to semantic labels. Both the source and target distributions share the same $K$ semantic class labels, ${1,...,K}$. Specifically, $X$ represents building patches, and $Y$ contains label maps with two class labels, namely \textit{background} and \textit{crack} ($K=2$).  We have access to a set of labeled source samples $S = {(x_j^s, y_j^s)},j=1,2,\ldots n_s$ and unlabeled target samples $T = {x_j^t},j=1,2,\ldots n_t$, where $n_s$ and $n_t$ denote the total number of source and target samples, respectively. Our objective is to train a network using the labeled source domain data $S$ and the unlabeled target domain data $T$ to generate accurate predictions ${\hat{y_j^t}},j=1,2,\ldots n_t$. In the context of crack segmentation, this problem is reduced to a binary segmentation task. However, due to the relatively small number of crack pixels present in each patch, a significant class imbalance exists.

\subsection{Incremental Learning} 

Incremental learning involves training an existing model on a sequence of $\tau$ tasks, where each task $\tau_i$ corresponds to a distinct dataset of domains $D_i$. In our specific setting, $D_i$ represents an image dataset consisting of pairs of input images and their corresponding semantic labels, denoted as $D_i = \{X_j, Y_j\}$. We will use $\tau$ and $D$ interchangeably. Each task $\tau_i$ is focused on semantic segmentation. Typically, a domain shift exists between consecutive tasks (i.e. $D_t$ exhibits non-trivial differences compared to $D_{t-1}$). The objective is to train a single semantic segmentation model $M$ that can effectively segment image data across each domain $D_t$ in a sequential manner. Thus, for a given task $\tau$, at each step $t$, our aim is to learn a mapping $M_t(X_t, t) = Y_t$ for the $t^{th}$ domain $D_t = (X_t, Y_t)$. Importantly, the learned model should maintain satisfactory performance on previous domains $D_{t-i}$, where $0 < i < t$, ensuring minimal degradation in performance.
\begin{figure*}[t]
    \centering
    \includegraphics[height=5cm,width=0.9\linewidth]{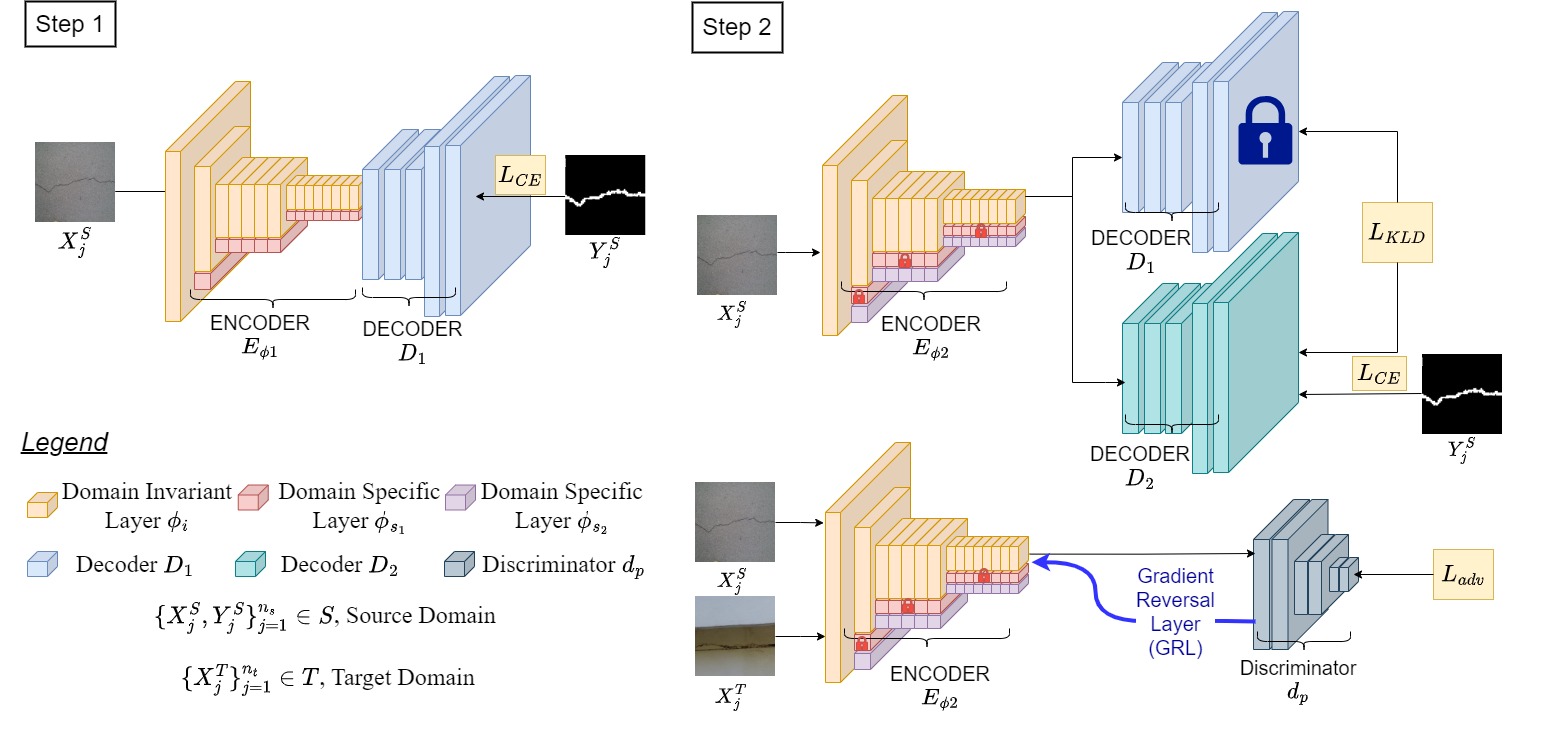}
    \caption{Overview of our proposed architecture (Section \ref{sec:implmentation}). In step 1 we train our network, $M_1$, using the labeled source dataset $S$ for binary segmentation. In step 2, decoder $D_1$ and $\phi_{s_1}$ are frozen, and a new set of domain-specific parameters $\phi_{s_2}$ are added and we call this model $M_2$. An alternating training strategy is followed in which we first train for binary segmentation on the source domain followed by adversarial training on both source and target domains.} 
    \label{fig:overviewmodel}
\end{figure*}
\subsection{Unsupervised Domain Adaptation (UDA)} 
UDA methods for semantic segmentation can be broadly classified into three groups: Self-training, Feature Alignment. and Adversarial Training approaches. Self-training approaches involve training a segmentation model on the labeled source domain to compute pseudo-labels \cite{psuedolabels} for the target domain. These pseudo-labels can be pre-computed offline \cite{crst} or online during training. To avoid training instabilities, several methods such as consistency regularization \cite{regularization1} based on data augmentation \cite{aug1}, domain mix-up \cite{dacs}, and pseudo-label prototypes \cite{protoype} have been used. Several methods also combine \cite{fada} self-training and adversarial training to perform UDA. Feature alignment \cite{featurealignment} approaches aim to align the feature representations of the source and target domains. This technique involves training a segmentation model with a domain adaptation loss, which encourages the feature representations of the source and target domains to be similar. For further details about UDA, we recommend reading \cite{udasurvey}. In the context of this work, we mainly focus on adversarial training. 

Adversarial training aims to minimize the domain discrepancy between the source and target distributions in a GAN framework \cite{grl}. The distributions can be aligned at the input \cite{advinput1}, output \cite{advent}, or patch \cite{advpatch} levels. For a GAN framework, the architecture (see Step 2 in Figure \ref{fig:overviewmodel}) is composed of a feature extractor ($E_{\phi_2}$), a label predictor ($D_1$ and $D_2$), a domain classifier ($d_\rho$), and a gradient reversal layer (GRL) in between $E_{\phi_2}$ and $d_\rho$. $d_\rho$ is trained to classify the source and target domains, while the segmentation model is trained to generate segmentation maps that are domain indistinguishable. A high-dimensional feature vector $x$ corresponding to input $X$ can be obtained below.
\begin{equation}
\label{eq:featurevector}
    E_\phi(X) = x
\end{equation}
In forward propagation, the GRL is implemented as an identity-mapping function while in back-propagation the GRL multiples the gradient calculated from the domain-classification error by a negative scalar. This negative gradient is propagated to the feature extractor. It can be formulated as below. 
\begin{gather}
    R_{\lambda} (x) =  x \\
    \frac{dR_{\lambda}}{dx} = - \lambda I
\end{gather}
$x$ is the corresponding feature vector for input $X$ obtained from Equation \ref{eq:featurevector}, $I$ is an identity matrix and $R_{\lambda}$ is the GRL. To mitigate the impact of large domain classification errors at the early stages of training, the value of $\lambda$ is regulated adaptively as given below where $p$ stands for the number of elapsed epochs. 
\begin{equation}
    \lambda = \frac{2}{1+e^{-\lambda p}} -1
    \label{eq:lambdaregulation}
\end{equation}
\section{Methodology}
\label{sec:methodology}

\subsection{Proposed Framework (CrackUDA)}

We design CrackUDA, a two-step unsupervised domain adaptation approach for binary segmentation of cracks (see Figure \ref{fig:overviewmodel}). Our model $M$ comprises an encoder $E_{\phi{_k}}$, two domain-specific decoders $D_{1}$ and $D_{2}$ for predicting domain-specific labels, and a discriminator network $d_\rho$ which acts as a domain classifier. The encoder $E_{\phi_k}$ consists of a set of shared domain-invariant parameters $\phi_i$  which are universal to all domains and a set of domain-specific parameters $\phi_{s_k}$ which are exclusive to respective domains. Domain-invariant parameters learn common features across all domains and domain-specific parameters learn domain-specific features for the respective domains.

As shown in Figure \ref{fig:overviewmodel}, the first step involves learning a binary segmentation $M_1$ on the source dataset $S$. $M_1$ is composed of decoder $D_{1}$ and encoder $E_{\phi_1}$ in which both $\phi_i$ and $\phi_{s_1}$ (domain-specific parameters for source dataset) are trainable. In step 2, we add new domain-specific parameters, $\phi_{s_2}$, to the new encoder $E_{\phi_2}$ and a domain-specific decoder $D_2$ and call this model $M_2$. We follow an alternating training strategy in which $M_2$ is trained for binary segmentation followed by adversarial training through the discriminator $d_\rho$. This training strategy enables our model to adapt to $T$ while retaining its performance on $S$. 

\begin{table}  
\scriptsize
  \centering
  \begin{tabular}{lccl}
    \toprule
     \textbf{Dataset} & \textbf{Size} & \textbf{\% of Crack} & \textbf{Description}\\
    \midrule    
    Crack500 \cite{crack500} & 3126& 6.03 & Collected using a smartphone\\
    Rissbilder \cite{rissbilder}  & 2736& 2.70 & Architectural Cracks\\
    SDNET2018 \cite{sdnet} & 1411 & 0 & Non-crack images \\
     Volker \cite{volker2015crack}  & 427& 4.05 & Cracks collected from pavements and buildings. \\
    DeepCrack \cite{deepcrack}  & 443& 3.58 & Cracks collected from pavements and buildings.\\
    GAPS384 \cite{gaps}  & 383& 1.21 & Cracks collected from pavements\\
    \textbf{BuildCrack (ours)} & \textbf{358}& \textbf{4.30} & \textbf{Building Cracks collected using a drone.}\\
    Masonry \cite{masonry}&  240& 4.21 & Contains crack in masonry walls\\
     CrackTree200 \cite{cracktree}  & 175& 0.31 & Cracks collected from \newline pavements and buildings.\\
    CFD \cite{cfd} & 118& 1.61 & Urban road surface cracks\\
    Ceramic \cite{ceramic}  & 100& 2.05 & Cracks on different colors and textures of ceramics.\\
    \bottomrule
\end{tabular}
\caption{Quantitative comparison of existing datasets and our dataset. The datasets mentioned above (except our new dataset) have been aggregated as CrackSeg9K~\cite{crackseg9k}.} 
  \label{tab:datasetdetails}
\end{table}
\subsection{Optimization Strategy}
For any given step $k$, the domain-specific layers $\phi_{s_k}$ are trained only on the softmax cross-entropy loss function over the label space of the source domain $S$. The forward pass and the softmax cross-entropy loss function, $\zeta$, can be formulated as given below.

\begin{gather}
    D_k(E_{\phi_k}(X_j, \phi_i, \phi_{s_k})) = \hat{Y_j} \label{eq:step1forward} \\
    L_{CE} = \frac{1}{N} \sum_{X_j, Y_j \in S} \zeta(Y_j, \hat{Y_j}) \label{eq:step1backward}
\end{gather}

 In step 2, in addition to the cross-entropy loss, we use a regularization loss $L_{KLD}$ to optimize the shared weights $\phi_i$ during the segmentation phase as given in the equations below.
 
\begin{gather}
     \hat{y}_j^{1} = M_1(X_j, \phi_i, \phi_{s_1}) \\
      \hat{y}_j^{2} = M_2(X_j, \phi_i, \phi_{s_1}) \\ 
     L_{KLD} = \sum_{X_j \in S} \psi (\hat{y}_j^{2}, \hat{y}_j^{1})
 \end{gather}

 where $\hat{y}_j^{1}$ and $\hat{y}_j^{2}$ are the softmax probability distributions maps of $M_1$ and $M_2$ on samples from the source domain respectively and $\psi$ is the KL-divergence loss between the two probability distributions. The total loss for the segmentation phase is given as below. 
 \begin{equation}
     L_{Total} = \lambda_{CE} \cdot L_{CE} + \lambda_{KLD} \cdot L_{KLD}
 \end{equation}
 
 For the adversarial training phase, we use a binary cross-entropy loss, $L_{adv}$ to classify whether the feature vector obtained from $E_{\phi_2}$ corresponds to an image sample from the source or target domain. This loss function can be formulated as given in the equations below. 
 \begin{gather}
     d_{\rho}(E_{\phi{_2}}(X_j, \phi_i, \phi_{s_2})) = \hat{d_j} \\ 
         L_{BCE} = \frac{1}{N} \sum_{X_j \in S, T} \omega(d_j, \hat{d_j})
 \end{gather}
 where $\omega$ is the binary cross-entropy loss, $d_j$ and $\hat{d_j}$ are the true and predicted domain labels and $d_{\rho}$ is the discriminator network. $d_j$ is a binary variable that indicates whether the sample is from the source or the target domain. 

\section{Implementation Details}

\label{sec:implmentation}

\subsection{Network Architecture} We use ERFNet \cite{erfnet} as the backbone for our network with the discriminator as an FCN. The value of $\lambda$ is updated as per Equation \ref{eq:lambdaregulation}. The encoder comprises residual-adapter blocks \cite{dau}. Each residual-adapter block has a set of domain-invariant parameters ($\phi_i = \{\phi_{w1}, \phi_{w2}\}$) and a set of domain-specific parameters ($\phi_{s_k} = \{\alpha^w, \alpha^s, \alpha^b \}$). $\phi_{w1}$ and $\phi_{w2}$ are $3 \times 3$ convolutional layers of a residual unit shared across all the domains. Domain-specific layers in the residual adapter unit are of two kinds: Domain-specific parallel residual adapter layers (DS-RAP) and domain-specific batch normalization layers (DS-BN). DS-RAP ($\alpha_w$) are $1 \times 1$ convolutional layers added to the shared convolutional layers in parallel. DS-BN shifts and scales the normalized input as $s \cdot x + b$ where $\alpha_s$ and $\alpha_b$ represent the scaling and shifting parameters respectively.

\subsection{Training}
In step 1 (see Figure \ref{fig:overviewmodel}), we train $M_1$ on the source domain $S$ in a binary segmentation setting. In step 2, we follow an alternating training strategy. We first train $M_2$ for binary segmentation for 10 epochs on $S$ followed by adversarial training on a mini-batch of an equal number of samples from $S$ and $T$ for 5 epochs. Overall, $M_2$ is trained for 150 epochs. $\lambda_{CE} $ and $\lambda_{KLD}$ are set to 1 and 0.1 respectively. In Step 1, segmentation is performed on $S$ for a total of 150 epochs. The Adam optimizer is utilized with a learning rate (LR) of $5e^{-4}$, and a batch size of $8$. In Step 2, segmentation is again performed on $S$ for 10 epochs, employing the same optimizer, learning rate, and batch size as in Step 1. Additionally, an adversarial training step is introduced, involving both the source ($S$) and target ($T$) datasets. This adversarial training step is conducted for 5 epochs. Training protocols have been summarized in Algorithm 1 and 2 in the supplementary material. For both steps, the model checkpoints were saved during training. For Step 2, The model checkpoints are saved only if there is an increase in mIoU scores for both the source and target domains.

\section{Experiments and Results} \label{sec:results}
\begin{table}
\resizebox{\linewidth}{!}
{
    \centering
    \begin{tabular}{|p{2.0cm}|p{1.5cm}|p{1.5cm}|p{1.5cm}|p{1.5cm}|p{1.5cm}|p{1.5cm}|p{1.5cm}|p{1.5cm}|}
\hline
\multirowcell{3}[0pt][l]{Dataset \\Excluded} & \multicolumn{4}{c|}{Step 1} & %
    \multicolumn{4}{c|}{Step 2} \\
    \cline{2-9}
    & \multirowcell{2}[0pt][l]{Source \\ mIoU} & \multicolumn{3}{c|}{Target mIoU} & \multirowcell{2}[0pt][l]{Source \\ mIoU} & \multicolumn{3}{c|}{Target mIoU} \\
    \cline{3-5} \cline{7-9}
    & & Dataset Excluded & \textbf{Build Crack}  & Overall (Excluded + Our) & & Dataset Excluded   & \textbf{Build Crack} & Overall (Excluded + Our)  \\
    \hline  
    Mason & 82.72 & 53.03 & 54.69 & 54.12 & 79.94 & \textbf{61.94} & \textbf{55.35} & \textbf{57.62}  \\
    \hline
    Ceramic & 82.67 & 49.55 & 62.55 & 59.98 & 78.86 & \textbf{50.55} & \textbf{63.73} & \textbf{62.16}  \\
    \hline
    CFD & 82.87 & 78.83 & 62.57 & 67.92 & 79.91 & \textbf{79.08} & 55.91 & 63.80  \\
    \hline
    Crack500 & 83.30 & 56.84 & 62.58 & 57.27 & 78.33 & \textbf{79.24} & 54.16 & \textbf{78.10}  \\
    \hline
    CrackTree200 & 82.52 & 77.64 & 57.69 & 66.61 & 79.26 & \textbf{81.48} & 52.05 & 65.28  \\
    \hline
    DeepCrack & 82.24 & 78.92 & 59.61 & 72.78 & 78.61 & \textbf{82.55} & 59.02 & \textbf{74.71}  \\
    \hline
    GAPS & 82.77 & 65.03 & 60.37 & 62.71 & 78.47 & \textbf{70.62} & 59.57 & \textbf{65.26}  \\
    \hline
    Rissbilder & 82.90 & 71.92 & 57.02 & 70.19 & 79.97 & \textbf{78.33} & 55.36 & \textbf{75.40}  \\
    \hline
    Volker & 82.64 & 75.20 & 57.80 & 69.98 & 79.77 & \textbf{76.80} & 57.60 & \textbf{70.40}  \\
    \hline

\end{tabular}
}
\caption{mIoU scores of CrackUDA (our approach) for steps 1 and 2 for sub-datasets of CrackSeg9K and BuildCrack (our custom dataset). Here, \textit{Dataset Excluded} is the sub-dataset left out of training and validation sets of the source domain. This \textit{Dataset Excluded} is aggregated with our dataset to form the overall dataset. Source mIoU is the performance of the network on the CrackSeg9K validation set excluding the mentioned dataset. The results show that using an incremental learning strategy for UDA leads to better performance in the target domain (see columns Dataset Excluded, Our Dataset, and Overall for Step 2) without a severe drop in performance in the source domain.}
\label{tab:subdatasetperformance}

\end{table}

\begin{figure*}[t]
    \centering
    \subfigure
    {
        \includegraphics[width=0.45\linewidth ]{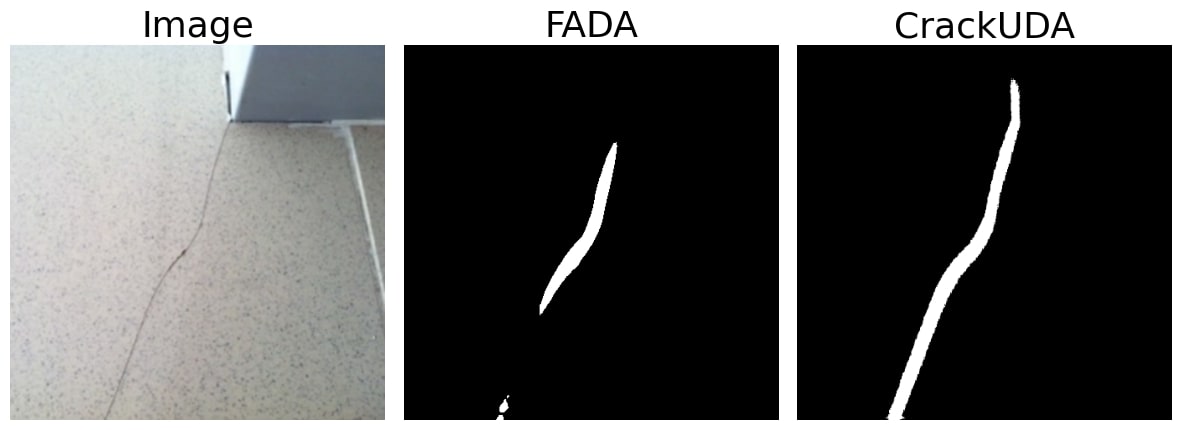}
        \label{fig:source_result1}
    }
    {
        \includegraphics[width=0.45\linewidth ]{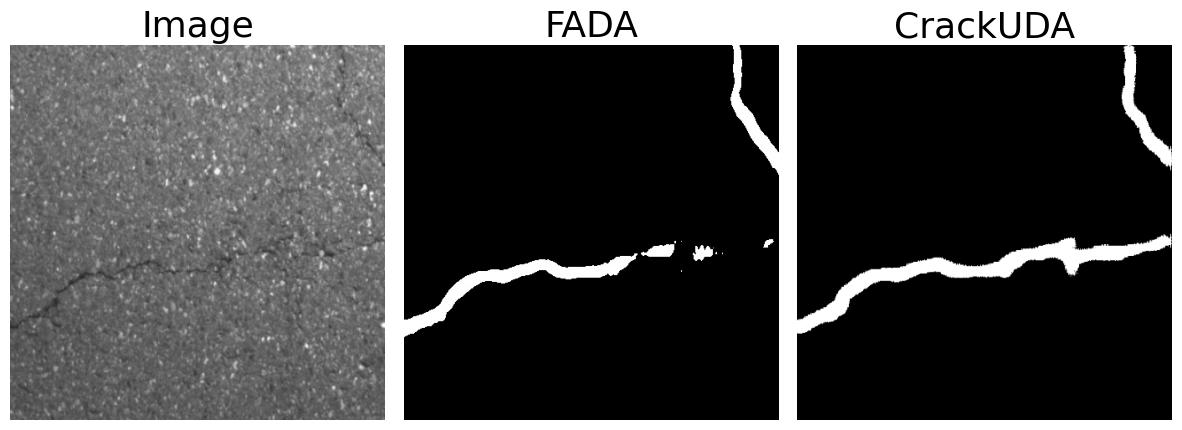}
        \label{fig:source_result2}
    }
    {
        \includegraphics[width=0.45\linewidth ]{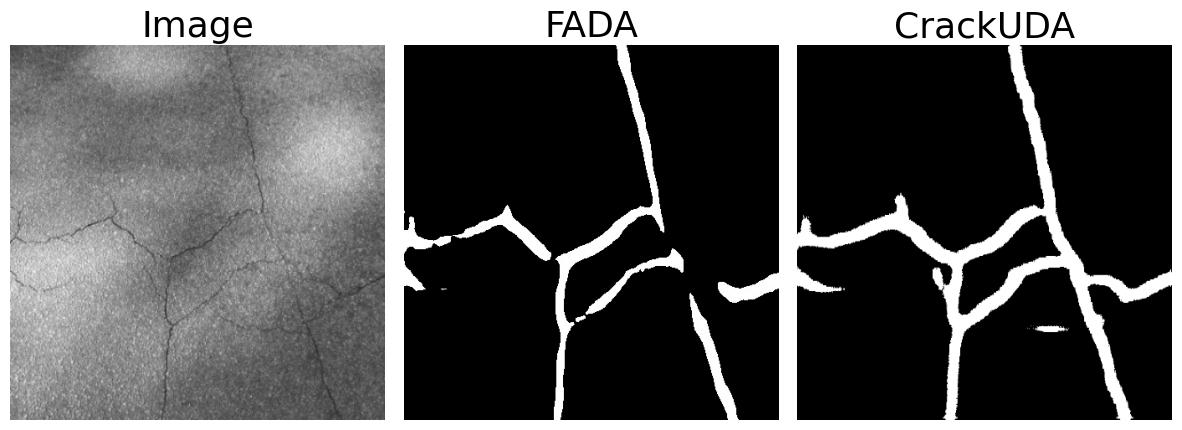}
        \label{fig:source_result3}
    }
    {
        \includegraphics[width=0.45\linewidth ]{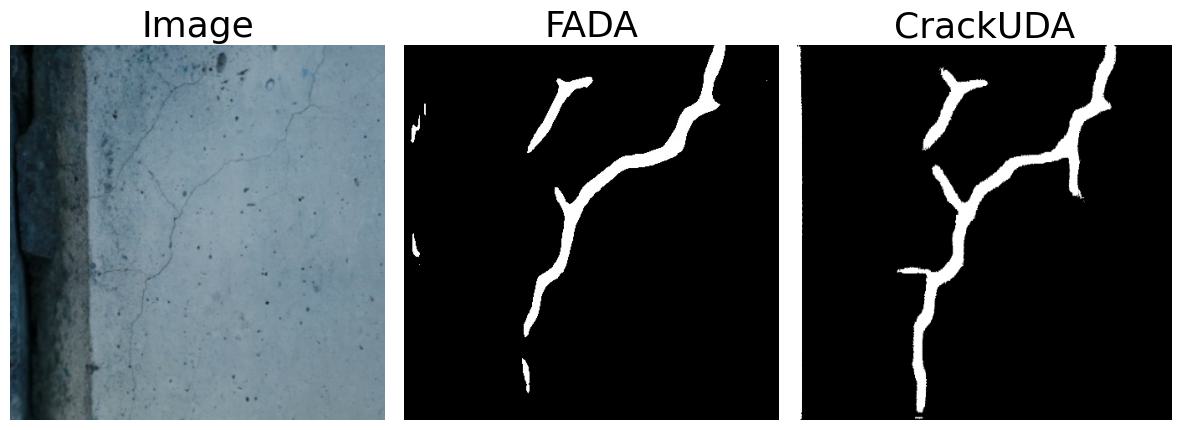}
        \label{fig:source_result4}
    }

    \caption{Qualitative results for CrackSeg9K validation set for CrackUDA and FADA \cite{fada}.}
        \label{fig:qualitativeresultscrack}
\end{figure*}
\begin{figure*}[t]
    \centering
    \subfigure
    {
        \includegraphics[width=0.45\linewidth]{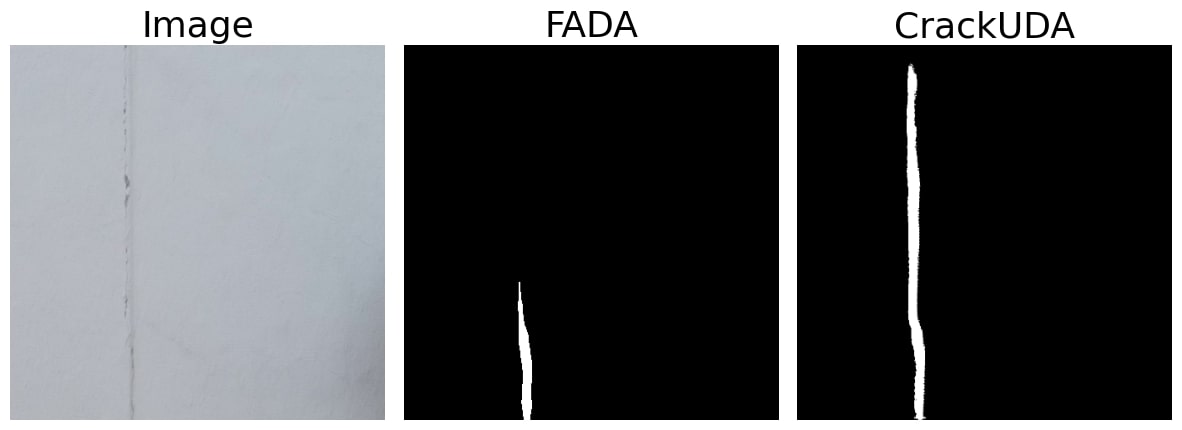}
        \label{fig:target_results1}
    }
    \subfigure
    {
        \includegraphics[width=0.45\linewidth]{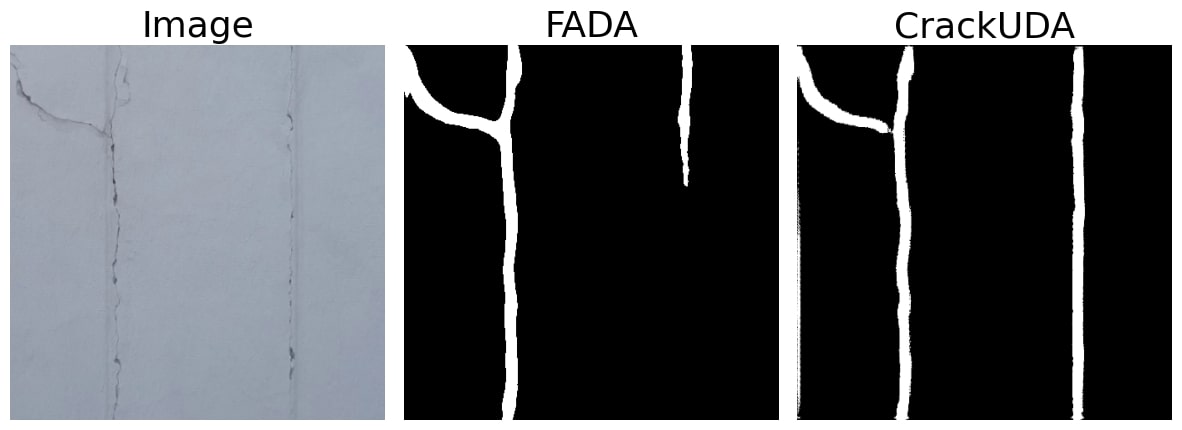}
        \label{fig:target_results2}
    }
    \subfigure
    {
        \includegraphics[width=0.45\linewidth]{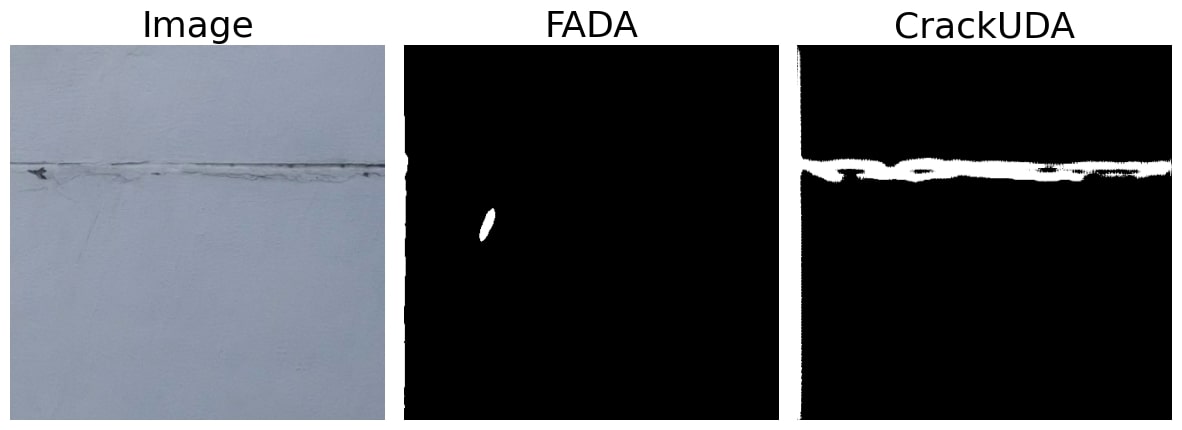}
        \label{fig:target_results3}
    }
    \subfigure
    {
        \includegraphics[width=0.45\linewidth]{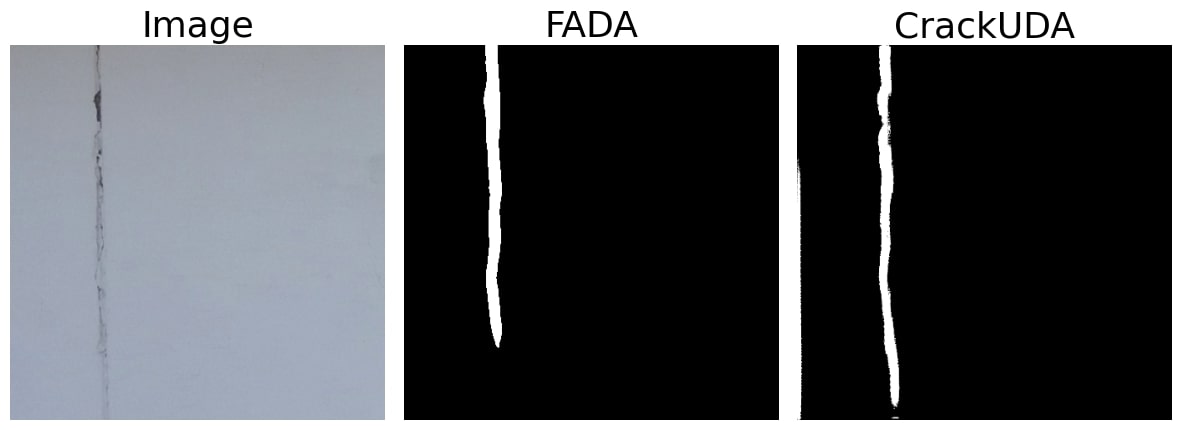}
        \label{fig:target_results4}
    }

    \caption{Qualitative results for BuildCrack for our network and FADA \cite{fada}.}
    \label{fig:qualitativeresultstarget}
\end{figure*}

\begin{figure*}[t]
    \centering
    \subfigure
    {
        \includegraphics[width=0.45\linewidth]{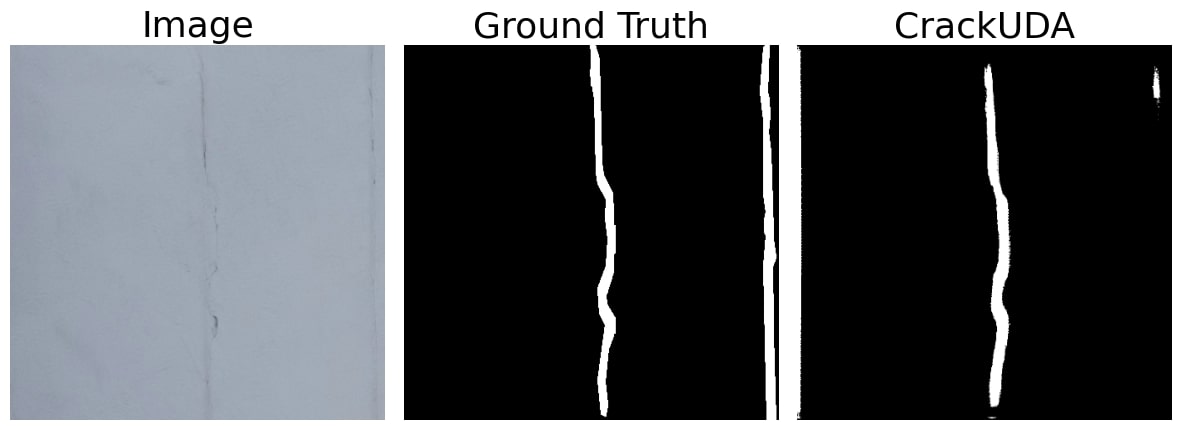}
        \label{fig:failure_results1}
    }
    \subfigure
    {
        \includegraphics[width=0.45\linewidth]{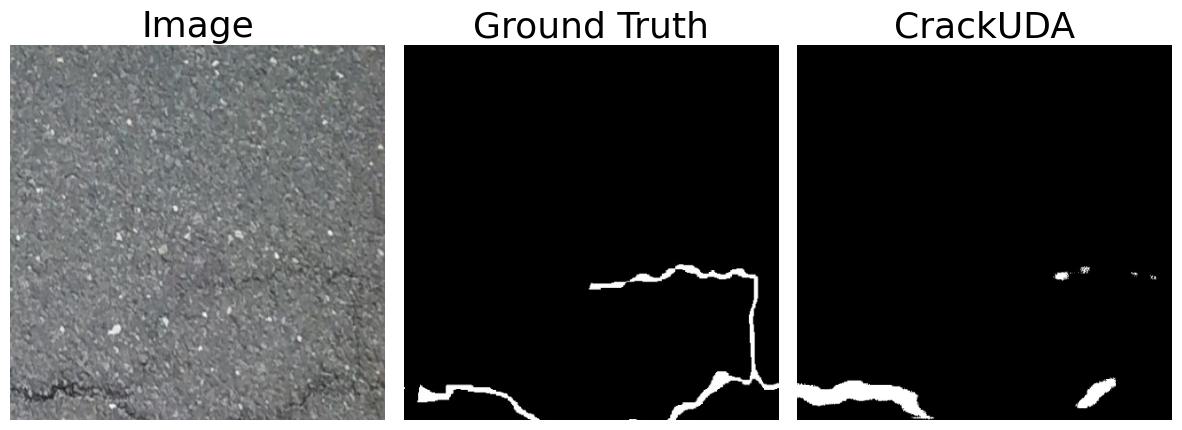}
        \label{fig:failure_results2}
    }
    \subfigure
    {
        \includegraphics[width=0.45\linewidth]{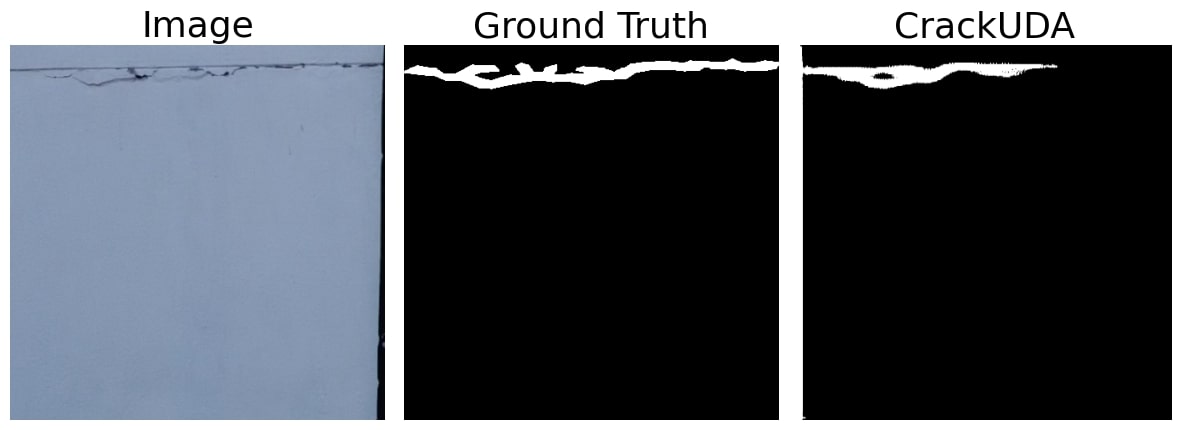}
        \label{fig:failure_results3}
    }
    \subfigure
    {
        \includegraphics[width=0.45\linewidth]{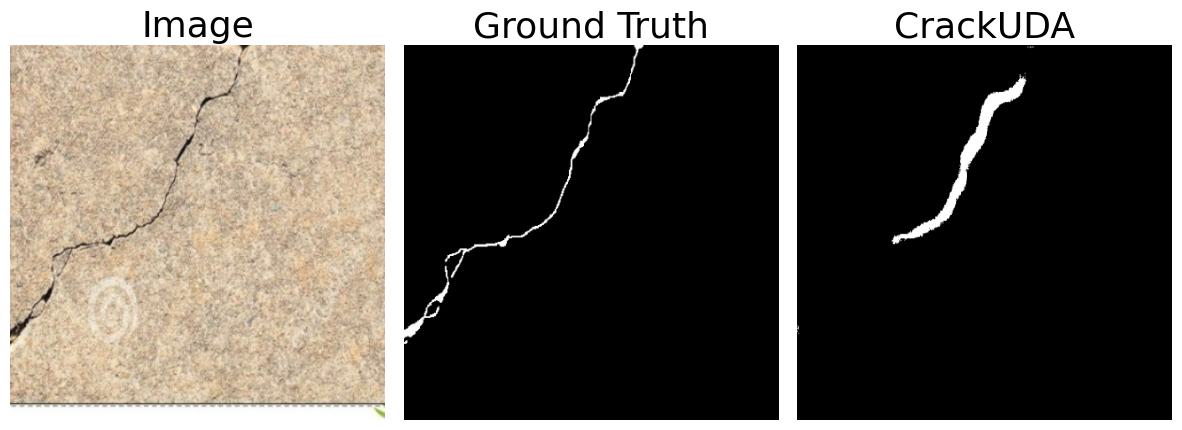}
        \label{fig:failure_results4}
    }

    \caption{Some cases in which our approach does not perform well in CrackSeg9K and BuildCrack.}
    \label{fig:failureresults}
\end{figure*}

\begin{table}[!t]
\centering
\scriptsize
\caption{Comparison of mIoU scores on the validation set of CrackSeg9K and BuildCrack (target dataset) with state-of-the-art UDA methods. \textsuperscript{*} approaches did not converge for our setting. Our approach achieves the best generalization performance.} 
  \label{tab:udabaseline}
  \begin{tabular}{ccc}
    \toprule
    DA Method & Source (\textbf{CrackSeg9k}) & Target (\textbf{BuildCrack}) \\
    \midrule
    AdaptSegnet\textsuperscript{*} \cite{adaptsegnet}& 47.53 & 48.47\\
    MaxSquare (ICCV '19) \cite{maxsquare}& 57.60& 50.50 \\
    ADVENT\textsuperscript{*} (CVPR '19) \cite{advent}& 47.51 & 48.47\\
    IAST\textsuperscript{*} (ECCV '20) \cite{iast}& 46.79& 46.78\\
    DAFormer\textsuperscript{*} (CVPR '22) \cite{daformer}& 47.54 & 48.47\\

    DACS (WACV '21) \cite{dacs}& 58.46 & 58.11\\

    CBST \textsuperscript{*} (ECCV '18) \cite{cbst} & 47.53 & 48.47 \\

    ProDA (CVPR '21) \cite{proda}  & 50.32 & 47.94 \\

    FADA (ECCV '20) \cite{fada}  &79.18 & 60.73\\
    \textbf{CrackUDA} & \textbf{79.83} & \textbf{63.43}\\
    \bottomrule
  
\end{tabular}
\end{table}
\subsection{Datasets and Evaluation Metrics} We validate the performance of our approach using two datasets: CrackSeg9K \cite{crackseg9k} and BuildCrack, the custom dataset that we introduce (see Figure \ref{fig:iiitdataset}). BuildCrack has images with low contrast, occlusions, and shadows, which challenge the model's robustness. CrackSeg9K is a culmination of smaller open-source crack datasets (CFD \cite{cfd}, Masonry \cite{masonry}, Ceramic \cite{ceramic}, Rissbilder \cite{rissbilder}, Volker \cite{volker2015crack}, SDNET2018 \cite{sdnet}, DeepCrack \cite{deepcrack}, GAPS384 \cite{gaps}, Crack500 \cite{crack500}, and CrackTree200 \cite{cracktree}) with more consistent labeling. Details regarding these 10 sub-datasets can be found in Table \ref{tab:datasetdetails}. After removing duplicate images in the original dataset of 9255 images, we divided the remaining 8513 images into 6794 training images, and 1719 validation images. This ratio of 4:1 was maintained across all sub-datasets to ensure an equal proportion of each sub-dataset in both the training and validation sets. Our dataset BuildCrack comprises 358 binary labeled crack images collected using DJI Mavic Mini\footnote{UAV specification details can be found at the official DJI website: \href{https://www.dji.com/mavic-mini}{https://www.dji.com/mavic-mini}}. All the ground-truth labels in both CrackSeg9K and BuildCrack have two class labels: \textit{background} and \textit{crack}. We use all 358 BuildCrack images for training and validation. We use mean Intersection-over-Union (mIoU) to evaluate the performance of our approach.

\subsection{UDA Baselines} We evaluate the performance of our network (CrackUDA) against 8 state-of-the-art UDA baselines and a state-of-the-art self-supervised UDA baseline in which CrackSeg9K and our dataset (BuildCrack) are the source and target datasets respectively. The performance of our approach and the baselines are reported on the validation set of CrackSeg9K and all 358 images of the BuildCrack (see Table \ref{tab:udabaseline}). \cite{advent, iast, cbst, daformer} did not converge for this setting. Out of the baselines, FADA \cite{fada} obtains the best mIoU score of 79.18 on the validation set of CrackSeg9K and 60.73 on our dataset. CrackUDA outperformed FADA by 0.65 and 2.7 mIoU on the validation set of CrackSeg9K and the entire BuildCrack dataset respectively.

\subsection{Experiments on sub-datasets of CrackSeg9K} 
We conduct experiments on sub-datasets of CrackSeg9K, systematically excluding one sub-dataset at a time from both the training and validation sets of the source domain $S$ during the two-step process. This exclusion preserves the 4:1 ratio between training and validation sets, maintaining the proportion of samples within each subset. The excluded dataset, combined with BuildCrack, forms the target dataset $T$. Table \ref{tab:subdatasetperformance} presents the mIoU scores obtained on the source dataset, excluded dataset, BuildCrack, and the new target dataset (excluded dataset + BuildCrack). The results show a significant increase in mIoU scores for the excluded datasets in Step 2. We observe an mIoU increase of 8.91 for Masonry, 0.99 for Ceramic, 0.25 for CFD, 22.4 for Crack500, 3.84 for CrackTree200, 3.63 for DeepCrack, 5.59 for GAPS, 6.41 for Rissbilder, and 1.60 for Volker. This demonstrates the generalization capabilities of our approach across target domains without notable decline in performance on the source domain. A comparison of our approach against a state-of-the-art supervised approach and performance impact due to switching source and target domains can be found in the supplementary material.

\subsection{Ablation Studies} In step 2 of our approach, we use $L_{KLD}$ to optimize the shared parameters $\phi_i$ through the softmax probability maps obtained from the domain-specific decoders. Our experiments show that removing this loss from step 2 leads to a 9.93 mIoU drop for the target dataset and a 0.84 mIoU drop for the source dataset (`2 Step w/o KLD' in Table \ref{tab:ablation}). This shows that optimizing for the shared parameters $\phi_i$ in step 2 helps the network learn common features of the source and target domain leading to better generalization across both domains. Next, we show that disabling adversarial training in step 2 leads to a 0.84 mIoU drop in the source dataset and a 1.61 mIoU drop in the target dataset (referred to as 2 Step w/o GRL in Table \ref{tab:ablation}). Intuitively, GRL plays a significant role in adapting the network to unlabeled target data. Overall, these ablation studies indicate that our proposed network with GRL and $L_{KLD}$ leads to the best overall performance on both the source and target domains. Analysis of the impact of $\lambda_{CE}$ and $\lambda_{KLD}$ can be found in the supplementary material.
\begin{table}[!t]

  \centering
  \newcolumntype{P}[1]{>{\centering\arraybackslash}p{#1}}
  \begin{tabular}{P{3cm}|P{1cm}|P{1cm}|P{2cm}|P{2cm}}
    \hline
     Method &$L_{KLD}$ &GRL &CrackSeg9K   & \textbf{BuildCrack} \\
\hline    1 Step & $\times$ & $\times$  & 82.17 & 60.44\\
    2 Step w/o GRL& \checkmark & $\times$  & 78.99 & 61.82\\
    2 Step w/o KLD & $\times$ & \checkmark  & 78.99 & 53.5\\
    2 Step & \checkmark & \checkmark  & 79.83  & 63.43 \\
    \hline
\end{tabular}

\caption{Ablation study on the contribution of each component of CrackUDA for the validation set of CrackSeg9K (source domain) and BuildCrack (target domain) setting.} 
\label{tab:ablation}
\end{table}
\section{Conclusion}
We propose CrackUDA, a novel two-step incremental Unsupervised Domain Adaptation (UDA) approach to address the challenging task of crack segmentation in civil structures. Our approach stands out from existing UDA methods by effectively addressing the issue of catastrophic forgetting through simultaneous learning of domain-invariant and domain-specific representations. Our experimental results demonstrate notable improvements, with 0.65 mIoU and 2.7 mIoU improvement on the source and target domains. Furthermore, we showcase the generalization capabilities of our approach across various sub-datasets of CrackSeg9K, and BuildCrack, our custom-created dataset. By providing an effective solution through incremental UDA, our work makes significant contributions to crack localization and structural health assessment in civil engineering. Additionally, our approach could serve as a benchmark to the research community focusing on unsupervised domain adaptation for semantic segmentation. 
\paragraph{\textbf{Acknowledgement:}}The authors acknowledge the financial support provided by IHUB, IIIT Hyderabad to carry out this research work under the project: IIIT-H/IHub/Project/Mobility/2021-22/M2-003.


%
%
%
\bibliographystyle{splncs04}
\bibliography{main}

\begin{thebibliography}{10}
\providecommand{\url}[1]{\texttt{#1}}
\providecommand{\urlprefix}{URL }
\providecommand{\doi}[1]{https://doi.org/#1}

\bibitem{rissbilder}
Bianchi, E., Hebdon, M.: {Concrete Crack Conglomerate Dataset}  (10 2021). \doi{10.7294/16625056.v1}

\bibitem{deeplab}
Chen, L.C., Zhu, Y., Papandreou, G., Schroff, F., Adam, H.: Encoder-decoder with atrous separable convolution for semantic image segmentation. In: Computer Vision -- ECCV 2018. pp. 833--851. Springer International Publishing, Cham (2018)

\bibitem{maxsquare}
Chen, M., Xue, H., Cai, D.: Domain adaptation for semantic segmentation with maximum squares loss. In: ICCV (October 2019)

\bibitem{od2}
Chen, X., Mottaghi, R., Liu, X., Fuchs, T., Yuille, A.: Unsupervised domain adaptation for object detection via back-propagation. In: Proceedings of the European Conference on Computer Vision. pp. 784--800 (2018)

\bibitem{od3}
Chen, Y., Li, W., Sakaridis, C., Dai, D., Van~Gool, L.: Domain adaptive faster r-cnn for object detection in the wild. In: Proceedings of the IEEE Conference on Computer Vision and Pattern Recognition. pp. 3339--3348 (2018)

\bibitem{curvilinear}
Cheng, M., Zhao, K., Guo, X., Xu, Y., Guo, J.: Joint topology-preserving and feature-refinement network for curvilinear structure segmentation. In: Proceedings of the IEEE/CVF International Conference on Computer Vision. pp. 7147--7156 (2021)

\bibitem{aug1}
Choi, J., Kim, T., Kim, C.: Self-ensembling with gan-based data augmentation for domain adaptation in semantic segmentation. In: ICCV 2019. pp. 6829--6839 (2019). \doi{10.1109/ICCV.2019.00693}

\bibitem{masonry}
Dais, D., İhsan Engin~Bal, Smyrou, E., Sarhosis, V.: Automatic crack classification and segmentation on masonry surfaces using convolutional neural networks and transfer learning. Automation in Construction  \textbf{125},  103606 (2021)

\bibitem{sdnet}
Dorafshan, S., Thomas, R.J., Maguire, M.: Sdnet2018: An annotated image dataset for non-contact concrete crack detection using deep convolutional neural networks. Data in Brief  \textbf{21},  1664--1668 (2018)

\bibitem{vit}
Dosovitskiy, A., Beyer, L., Kolesnikov, A., Weissenborn, D., Zhai, X., Unterthiner, T., Dehghani, M., Minderer, M., Heigold, G., Gelly, S., Uszkoreit, J., Houlsby, N.: An image is worth 16x16 words: Transformers for image recognition at scale. CoRR  \textbf{abs/2010.11929} (2020)

\bibitem{gaps}
Eisenbach, M., Stricker, R., Seichter, D., Amende, K., Debes, K., Sesselmann, M., Ebersbach, D., Stoeckert, U., Gross, H.M.: How to get pavement distress detection ready for deep learning? a systematic approach. In: IJCNN 2017. pp. 2039--2047 (2017). \doi{10.1109/IJCNN.2017.7966101}

\bibitem{classification2}
Ganin, Y., Lempitsky, V.: Unsupervised domain adaptation by backpropagation. In: International conference on machine learning. pp. 1180--1189 (2015)

\bibitem{grl}
Ganin, Y., Ustinova, E., Ajakan, H., Germain, P., Larochelle, H., Laviolette, F., Marchand, M., Lempitsky, V.: Domain-adversarial training of neural networks (2016)

\bibitem{dau}
Garg, P., Saluja, R., Balasubramanian, V.N., Arora, C., Subramanian, A., Jawahar, C.: Multi-domain incremental learning for semantic segmentation. In: Proceedings of the IEEE/CVF Winter Conference on Applications of Computer Vision. pp. 761--771 (2022)

\bibitem{advinput1}
Gong, R., Li, W., Chen, Y., Van~Gool, L.: Dlow: Domain flow for adaptation and generalization. In: 2019 IEEE/CVF Conference on Computer Vision and Pattern Recognition (CVPR). pp. 2472--2481 (2019)

\bibitem{featurealignment}
Hoffman, J., Wang, D., Yu, F., Darrell, T.: Fcns in the wild: Pixel-level adversarial and constraint-based adaptation. CoRR  \textbf{abs/1612.02649} (2016)

\bibitem{daformer}
Hoyer, L., Dai, D., Van~Gool, L.: {DAFormer}: Improving network architectures and training strategies for domain-adaptive semantic segmentation. In: Proceedings of the IEEE/CVF Conference on Computer Vision and Pattern Recognition (CVPR). pp. 9924--9935 (2022)

\bibitem{ceramic}
Junior, G.S., Ferreira, J., Millán-Arias, C., Daniel, R., Junior, A.C., Fernandes, B.J.T.: Ceramic cracks segmentation with deep learning. Applied Sciences  \textbf{11}(13) (2021). \doi{10.3390/app11136017}

\bibitem{cfd}
Khalesi, S., Ahmadi, A.: Automatic road crack detection and classification using image processing techniques, machine learning and integrated models in urban areas: A novel image binarization technique  (06 2020)

\bibitem{KOCH2015196}
Koch, C., Georgieva, K., Kasireddy, V., Akinci, B., Fieguth, P.: A review on computer vision based defect detection and condition assessment of concrete and asphalt civil infrastructure. Advanced Engineering Informatics  \textbf{29}(2),  196--210 (2015). \doi{https://doi.org/10.1016/j.aei.2015.01.008}, infrastructure Computer Vision

\bibitem{kondo2021crack_cssr}
Kondo, Y., Ukita, N.: Crack segmentation for low-resolution images using joint learning with super-resolution. In: 2021 17th International Conference on Machine Vision and Applications (MVA). pp.~1--6. IEEE (2021)

\bibitem{konig2022weakly_weaksup_seg}
K{\"o}nig, J., Jenkins, M.D., Mannion, M., Barrie, P., Morison, G.: Weakly-supervised surface crack segmentation by generating pseudo-labels using localization with a classifier and thresholding. IEEE Transactions on Intelligent Transportation Systems  \textbf{23}(12),  24083--24094 (2022)

\bibitem{crackseg9k}
Kulkarni, S., Singh, S., Balakrishnan, D., Sharma, S., Devunuri, S., Korlapati, S.C.R.: Crackseg9k: a collection and benchmark for crack segmentation datasets and frameworks. In: Computer Vision--ECCV 2022 Workshops: Tel Aviv, Israel, October 23--27, 2022, Proceedings, Part VII. pp. 179--195. Springer (2023)

\bibitem{lateef2019survey_segmentation}
Lateef, F., Ruichek, Y.: Survey on semantic segmentation using deep learning techniques. Neurocomputing  \textbf{338},  321--348 (2019)

\bibitem{lau2020automated}
Lau, S.L., Chong, E.K., Yang, X., Wang, X.: Automated pavement crack segmentation using u-net-based convolutional neural network. Ieee Access  \textbf{8},  114892--114899 (2020)

\bibitem{psuedolabels}
Lee, D.H.: Pseudo-label : The simple and efficient semi-supervised learning method for deep neural networks. ICML 2013 Workshop : Challenges in Representation Learning (WREPL)  (07 2013)

\bibitem{semi_crackseg}
Li, G., Wan, J., He, S., Liu, Q., Ma, B.: Semi-supervised semantic segmentation using adversarial learning for pavement crack detection. IEEE Access  \textbf{8},  51446--51459 (2020). \doi{10.1109/ACCESS.2020.2980086}

\bibitem{od5}
Liu, F., Li, X., Wang, H., Cheng, J.: Domain adaptive faster r-cnn via cross-domain marginal alignment. In: Proceedings of the IEEE Conference on Computer Vision and Pattern Recognition. pp. 12016--12025 (2020)

\bibitem{liu2021crackformer}
Liu, H., Miao, X., Mertz, C., Xu, C., Kong, H.: Crackformer: Transformer network for fine-grained crack detection. In: Proceedings of the IEEE/CVF International Conference on Computer Vision. pp. 3783--3792 (2021)

\bibitem{deepcrack}
Liu, Y., Yao, J., Lu, X., Xie, R., Li, L.: Deepcrack: A deep hierarchical feature learning architecture for crack segmentation. Neurocomputing  \textbf{338},  139--153 (2019)

\bibitem{classification3}
Long, M., Zhu, H., Wang, J., Jordan, M.I.: Deep adaptation networks: A more general robustification scheme for deep learning. IEEE transactions on pattern analysis and machine intelligence  \textbf{41}(9),  1956--1970 (2018)

\bibitem{forgetting}
McCloskey, M., Cohen, N.J.: Catastrophic interference in connectionist networks: The sequential learning problem. Psychology of Learning and Motivation, vol.~24, pp. 109--165. Academic Press (1989)

\bibitem{iast}
Mei, K., Zhu, C., Zou, J., Zhang, S.: Instance adaptive self-training for unsupervised domain adaptation  (2020)

\bibitem{oliveira2017road}
Oliveira, H., Correia, P.L.: Road surface crack detection: Improved segmentation with pixel-based refinement. In: 2017 25th European Signal Processing Conference (EUSIPCO). pp. 2026--2030. IEEE (2017)

\bibitem{intrada}
Pan, F., Shin, I., Rameau, F., Lee, S., Kweon, I.S.: Unsupervised intra-domain adaptation for semantic segmentation through self-supervision. In: IEEE Conference on Computer Vision and Pattern Recoginition (CVPR) (2020)

\bibitem{classification1}
Pei, W., Wang, Y., Vigneron, V., Wu, T.: Adversarial discriminative domain adaptation. In: Proceedings of the IEEE Conference on Computer Vision and Pattern Recognition. pp. 7167--7176 (2018)

\bibitem{primer}
Ramancharla, P., Bhalkikar, A., Velani, P., Vyas, P., Prakke, B., Patnala, N., Talyan, N.: A primer on rapid visual screening (rvs) consolidating earthquake safety assessment efforts in india  (10 2020)

\bibitem{erfnet}
Romera, E., Alvarez, J.M., Bergasa, L., Arroyo, R.: Erfnet: Efficient residual factorized convnet for real-time semantic segmentation. IEEE Transactions on Intelligent Transportation Systems  \textbf{PP},  1--10 (10 2017)

\bibitem{udasurvey}
Schwonberg, M., Niemeijer, J., Termöhlen, J.A., schäfer, J.P., Schmidt, N.M., Gottschalk, H., Fingscheidt, T.: Survey on unsupervised domain adaptation for semantic segmentation for visual perception in automated driving. IEEE Access  \textbf{11},  54296--54336 (2023). \doi{10.1109/ACCESS.2023.3277785}

\bibitem{regularization1}
Tarvainen, A., Valpola, H.: Weight-averaged consistency targets improve semi-supervised deep learning results. CoRR  \textbf{abs/1703.01780} (2017)

\bibitem{dacs}
Tranheden, W., Olsson, V., Pinto, J., Svensson, L.: Dacs: Domain adaptation via cross-domain mixed sampling. In: Proceedings of the IEEE/CVF winter conference on applications of computer vision. pp. 1379--1389 (2021)

\bibitem{adaptsegnet}
Tsai, Y.H., Hung, W.C., Schulter, S., Sohn, K., Yang, M.H., Chandraker, M.: Learning to adapt structured output space for semantic segmentation. In: 2018 IEEE/CVF Conference on Computer Vision and Pattern Recognition. pp. 7472--7481 (2018). \doi{10.1109/CVPR.2018.00780}

\bibitem{advpatch}
Tsai, Y.H., Sohn, K., Schulter, S., Chandraker, M.: Domain adaptation for structured output via discriminative patch representations. In: Proceedings of the IEEE/CVF international conference on computer vision. pp. 1456--1465 (2019)

\bibitem{classfication4}
Tzeng, E., Hoffman, J., Zhang, N., Saenko, K., Darrell, T.: Learning transferable features with deep adaptation networks. In: International conference on machine learning. pp. 2208--2217 (2017)

\bibitem{volker2015crack}
Volker, A., Pahlavan, L., Blacquiere, G.: Crack depth profiling using guided wave angle dependent reflectivity. In: AIP Conference Proceedings. vol.~1650, pp. 785--791. American Institute of Physics (2015)

\bibitem{advent}
Vu, T.H., Jain, H., Bucher, M., Cord, M., P{\'e}rez, P.: Advent: Adversarial entropy minimization for domain adaptation in semantic segmentation. In: Proceedings of the IEEE/CVF conference on computer vision and pattern recognition. pp. 2517--2526 (2019)

\bibitem{fada}
Wang, H., Shen, T., Zhang, W., Duan, L., Mei, T.: Classes matter: A fine-grained adversarial approach to cross-domain semantic segmentation. In: The European Conference on Computer Vision (ECCV) (August 2020)

\bibitem{weng2023unsupervised}
Weng, X., Huang, Y., Li, Y., Yang, H., Yu, S.: Unsupervised domain adaptation for crack detection. Automation in Construction  \textbf{153},  104939 (2023)

\bibitem{classification5}
Xu, J., Zhu, Y., Li, X., Li, C., Wang, M., Zhang, B., Lu, H.: Unsupervised domain adaptation with adversarial residual transform networks. In: Proceedings of the IEEE Conference on Computer Vision and Pattern Recognition. pp. 1365--1374 (2019)

\bibitem{crack500}
Yang, F., Zhang, L., Yu, S., Prokhorov, D., Mei, X., Ling, H.: Feature pyramid and hierarchical boosting network for pavement crack detection. IEEE Transactions on Intelligent Transportation Systems  \textbf{21}(4),  1525--1535 (2019)

\bibitem{zhang2017automated_cracknet}
Zhang, A., Wang, K.C., Li, B., Yang, E., Dai, X., Peng, Y., Fei, Y., Liu, Y., Li, J.Q., Chen, C.: Automated pixel-level pavement crack detection on 3d asphalt surfaces using a deep-learning network. Computer-Aided Civil and Infrastructure Engineering  \textbf{32}(10),  805--819 (2017)

\bibitem{protoype}
Zhang, P., Zhang, B., Zhang, T., Chen, D., Wang, Y., Wen, F.: Prototypical pseudo label denoising and target structure learning for domain adaptive semantic segmentation. In: Proceedings of the IEEE/CVF conference on computer vision and pattern recognition. pp. 12414--12424 (2021)

\bibitem{proda}
Zhang, P., Zhang, B., Zhang, T., Chen, D., Wang, Y., Wen, F.: Prototypical pseudo label denoising and target structure learning for domain adaptive semantic segmentation. arXiv preprint arXiv:2101.10979  (2021)

\bibitem{od4}
Zhang, Y., Qiao, Y., Liu, C., Shen, W., Wang, X.: Domain adaptive faster r-cnn with co-attention networks. In: Proceedings of the IEEE International Conference on Computer Vision. pp. 3695--3704 (2019)

\bibitem{cracktree}
Zou, Q., Cao, Y., Li, Q., Mao, Q., Wang, S.: Cracktree: Automatic crack detection from pavement images. Pattern Recognition Letters  \textbf{33}(3),  227--238 (2012)

\bibitem{cbst}
Zou, Y., Yu, Z., Kumar, B.V., Wang, J.: Unsupervised domain adaptation for semantic segmentation via class-balanced self-training. In: Proceedings of the European Conference on Computer Vision (ECCV). pp. 289--305 (2018)

\bibitem{crst}
Zou, Y., Yu, Z., Liu, X., Kumar, B., Wang, J.: Confidence regularized self-training. In: Proceedings of the IEEE/CVF international conference on computer vision. pp. 5982--5991 (2019)

\end{thebibliography}

\end{document}


\title{CrackUDA: Incremental Unsupervised Domain Adaptation for Improved Crack Segmentation in Civil Structures (Supplementary)}
%
\author{Kushagra Srivastava\and
Damodar Datta Kancharla \and
Rizvi Tahereen \and Pradeep Kumar Ramancharla \and Ravi Kiran Sarvadevabhatla \and Harikumar Kandath}
\authorrunning{K. Srivastava et al.}
%
\institute{International Institute of Information Technology, Hyderabad, India}
\maketitle

\section{Training Procedure}
\subsection{Training}
The training procedure described in Section 5.2 has been summarized in Algorithms \ref{alg:step1} and \ref{alg:step2}. 
\renewcommand{\algorithmicrequire}{\textbf{Input:}}
\renewcommand{\algorithmicensure}{\textbf{Output:}}
\begin{algorithm}
\caption{Training Protocol for Step 1}
\label{alg:step1}
\begin{algorithmic}
\Require \\
    $\{X_j, Y_j\}_{j=1}^{n_s}  \in S$\\
    $M$: Model composed of $E_{\phi_1}$ and $D_{1}$ 
\Ensure \\
Predicted softmax probability maps $\hat{Y}_j$ 

\For {epochs}{
    \For {mini-batch}{

        Forward pass: $D_{1}(E_{\phi_1}(X_j, \phi_{s_1}, \phi_i)) = \hat{Y_j}$ \\
         \hskip2.0em Compute CE Loss: $L_{CE}$ using $\hat{Y_j}$ and $Y_j$\\
        \hskip2.0em \textbf{Update:} $E_{\phi_1}$  and $D_1$ weights using $L_{CE}$
        }
        \EndFor
    }
\EndFor
\end{algorithmic}
\end{algorithm}
\renewcommand{\algorithmicrequire}{\textbf{Input:}}
\renewcommand{\algorithmicensure}{\textbf{Output:}}

\begin{algorithm}
\caption{Training Protocol for Step 2}
\label{alg:step2}
\begin{algorithmic}
\Require \\
    $\{X_j^S,Y_j^S \}_{j=1}^{n_s} \in S$; $\{X_j^{T}\}_{j=1}^{n_t} \in T$  \\
    $M$: Model composed of $E_{\phi_2}$, $D_{1}$, $D_{2}$ and $d_\rho$ \\
    Domain Classification labels: $d_i$ 
\Ensure \\
Segmentation Phase: Softmax probability maps $\hat{Y}_j$ \\
Adversarial Phase: Predicted domain labels $\hat{d}_j$ \\
\hskip-1.0em \textbf{Freeze:}\\
$\phi_{s_1}$ in $E_{\phi_2}$ and $D_1$ \\

\hskip-1.0em\textbf{Trainable:} \\

$E_{\phi_2}$ except $\phi_{s_1}$, $D_2$ and $d_\rho$
\For {epochs}{
    \For {m segmentation epochs}{
        \For{mini-batches}{      
         \hskip2.0em $ \{X_j^{S}, Y_j^{S}\} \sim S$ \\
        \hskip4.0em Forward pass: $D_{2}(E_{\phi_2}(X_j, \phi_{s_1}, \phi_i) = \hat{y}_j^2$ \\
        \hskip4.0em $D_{1}(E_{\phi_1}(X_j, \phi_{s_1}, \phi_i)) = \hat{y}_j^1$ \\
        \hskip4.0em $D_{2}(E_{\phi_2}(X_j, \phi_{s_2}, \phi_i)) = \hat{Y}_j$ \\
         \hskip4.0em Compute KLD Loss: $L_{KLD}$ using $\hat{y}_j^1$ and $\hat{y}_J^2$\\
         \hskip4.0em Compute CE Loss: $L_{CE}$ using $\hat{Y_j}$ and $Y_j$\\
        \hskip4.0em \textbf{Update:} $D_2$, $\phi_{S_2}$ using $L_{CE}$, $\phi_i$ using $L_{KLD}$  
        }\EndFor
        } \EndFor
    \For {n DA epochs}{
         \For{mini-batches}{
         \hskip4.0em $X_j \sim \{X_j^{S}, X_j^{T}\}$ \\
        \hskip6.0em Forward pass: $d_\rho(E_{\phi_2}(X_j, \phi_i, \phi_{s_2})) = \hat{d_j}$ \\
         \hskip6.0em BCE Loss: $L_{adv}$ using $\hat{d_j}$ and $d_j$\\
        \hskip6.0em \textbf{Update:} $E_{\phi_1}$ using $L_{adv}$
        }\EndFor}
        \EndFor}
\EndFor

\end{algorithmic}
\end{algorithm}
\section{Additional Experiments}
\subsection{Comparison with supervised approach}
We performed experiments to compare our approach with a state-of-the-art supervised semantic segmentation approach \cite{deeplab}. We trained \cite{deeplab} on the \textit{source dataset} using the default training hyperparameters and the network does not see the \textit{target dataset} through the training process. We report the performance of \cite{deeplab} as mIoU scores (see columns 2-5 of Table \ref{tab:subdatasetperformanceappendix}) on the validation set of the \textit{source dataset}, \textit{dataset excluded}, \textit{BuildCrack}, and the \textit{target dataset} (aggregation of \textit{dataset excluded} and \textit{BuildCrack}). We evaluate the performance of our approach against the supervised approach (see columns 6-9 of Table \ref{tab:subdatasetperformanceappendix}). Our approach enables better generalization across the target dataset without a significant drop in performance on the \textit{source dataset}. In particular, we achieved an increase of 4.77 mIoU on CFD, 19.00 mIoU on Crack500, 6.16 mIoU on CrackTree200, 1.43 mIoU on Deepcrack, 6.01 mIoU on GAPS384, 9.45 mIoU on Rissbilder, and 5.64 mIoU on Volker when the aforementioned datasets were excluded from the \textit{source dataset}. On the \textit{target dataset}, we achieve an increase of 2.72, 17.57, 0.25, 3.62, 8.18, and 3.96 mIoU scores when Ceramic, Crack500, Deepcrack, GAPS384, Rissbilder, and Volker sub-datasets were excluded from the \textit{source dataset} respectively.
\begin{table}
\resizebox{\linewidth}{!}{
 \begin{tabular}{|p{2.0cm}|p{1.5cm}|p{1.5cm}|p{1.5cm}|p{1.5cm}|p{1.5cm}|p{1.5cm}|p{1.5cm}|p{1.5cm}|}
\hline
\multirowcell{3}[0pt][l]{Dataset \\Excluded} & \multicolumn{4}{c|}{Supervised Approach using \cite{deeplab}} & %
    \multicolumn{4}{c|}{Our Approach (Step 2)} \\
    \cline{2-9}
    & \multirowcell{2}[0pt][l]{Source \\ mIoU} & \multicolumn{3}{c|}{Target mIoU} & \multirowcell{2}[0pt][l]{Source \\ mIoU} & \multicolumn{3}{c|}{Target mIoU} \\
    \cline{3-5} \cline{7-9}
    & & Dataset Excluded & \textbf{Build Crack}  & Overall (Excluded + Our) & & Dataset Excluded   & \textbf{Build Crack} & Overall (Excluded + Our)  \\
    \hline  
    Mason & \textbf{80.17} & \textbf{63.33} & \textbf{62.38} & \textbf{60.91} & 79.94 & 61.94 & 55.35 & 57.62  \\
    \hline
    Ceramic & \textbf{80.61} & \textbf{61.52} & 58.61 & 59.44 & 78.86 & 50.55 & \textbf{63.73} & \textbf{62.16}  \\
    \hline
    CFD & \textbf{80.47} & 74.31 & \textbf{58.82} & \textbf{64.04} & 79.91 & \textbf{79.08} & 55.91 & 63.80  \\
    \hline
    Crack500 & \textbf{80.05} & 60.24 & \textbf{63.68} & 60.53 & 78.33 & \textbf{79.24} & 54.16 & \textbf{78.10}  \\
    \hline
    CrackTree200 & \textbf{80.78} & 75.32 & \textbf{62.57} & \textbf{68.35} & 79.26 & \textbf{81.48} & 52.05 & 65.28  \\
    \hline
    DeepCrack & \textbf{80.14} & 81.12 & \textbf{60.64} & 74.46 & 78.61 & \textbf{82.55} & 59.02 & \textbf{74.71}  \\
    \hline
    GAPS & \textbf{80.37} & 64.61 & 57.81 & 61.64 & 78.47 & \textbf{70.62} & \textbf{59.57} & \textbf{65.26}  \\
    \hline
    Rissbilder & \textbf{81.32} & 68.88 & \textbf{55.65} & 67.22 & 79.97 & \textbf{78.33} & 55.36 & \textbf{75.40}  \\
    \hline
    Volker & \textbf{80.92} & 71.16 & 56.40 & 66.44 & 79.77 & \textbf{76.80} & \textbf{57.60} & \textbf{70.40}  \\
    \hline

\end{tabular}
}
\caption{mIoU scores of the supervised approach \cite{deeplab} and step 2 of our approach for sub-datasets of CrackSeg9K and our custom dataset. Here, \textit{Dataset Excluded} is the sub-dataset left out of training and validation sets of the source domain. This \textit{Dataset Excluded} is aggregated with our dataset to form the overall dataset. Source mIoU is the performance of the network on the CrackSeg9k validation set excluding the mentioned dataset. The results show that our approach leads to better performance in the target domain when compared to \cite{deeplab}, a state-of-the-art supervised segmentation approach.}
\label{tab:subdatasetperformanceappendix}
\end{table}
\subsection{Switching Target and Source Domains}
We performed experiments to observe the performance of our approach by switching the source and target domains. We report results CrackSeg9k as source and BuildCrack as target (see row 1 of Table \ref{tab:swtichingtargetsource}) as well as BuildCrack as source and CrackSeg9k as target (see row 2 of Table \ref{tab:swtichingtargetsource}). It is observed that the latter does not generalize well in the target domain. This is because BuildCrack has a considerably smaller number of samples (BuildCrack has 358 images and the train set of CrackSeg9k has 6794 images) when compared to CrackSeg9k. We use the images from the training set of CrackSeg9k in Step 2 of our approach. The hyperparameters are not changed across experiments and we follow the same training procedure as mentioned in Section 5 of the main paper.

\begin{table}[htbp]
\centering
  \begin{tabular}{p{5cm}p{2cm}p{2cm}}
    \toprule
    Experiment & Source (mIoU) & Target (mIoU) \\
    \midrule
    CrackSeg9K$\to$ BuildCrack & \textbf{79.83} & \textbf{63.43} \\
    BuildCrack$\to$ CrackSeg9k & 74.91 & 51.50 \\
    \bottomrule
\end{tabular}
\caption{We follow the notation of Source $\to$ Target for denoting the experiments. We report mIoU scores on the entire BuildCrack dataset and the validation set of CrackSeg9k. These results showcase that it is better to transfer from a larger source domain to a smaller target domain.} 
  \label{tab:swtichingtargetsource}

\end{table}
\subsection{Impact of $\lambda_{CE}$ and $\lambda_{KLD}$}

We performed experiments to determine the optimal values for $\lambda_{CE}$ and $\lambda_{KLD}$. We report results as mIoU scores for Step 2 of our approach in Table \ref{tab:lambdaanaylsis}. For this setting, we follow the same training procedure and experimental setup as mentioned in Sections 5 and 6 of the main paper.  Our experiments show that $\lambda_{CE}=1.0$ and $\lambda_{KLD}=0.1$ enables better generalization across both domains. 

\begin{table}[htbp]
\centering
  \begin{tabular}{p{2cm}p{1.5cm}p{1.5cm}|p{2cm}p{1.5cm}p{1.5cm}}
    \toprule
    $\lambda_{CE}=1$ & Source (mIoU) & Target (mIoU)  & $\lambda_{KLD}=0.1$ & Source (mIoU) & Target (mIoU)\\
    \midrule
    $\lambda_{KLD}=0.1$ & 79.83 & \textbf{63.43} & $\lambda_{CE}=1.0$ & \textbf{79.83} & \textbf{63.43}\\
    \midrule
    $\lambda_{KLD}=0.2$ & 78.97 & 62.59 & $\lambda_{CE}=0.9$ & 78.80 & 61.20\\
    \midrule
    $\lambda_{KLD}=0.3$ & 78.47 & 61.57 &     $\lambda_{CE}=0.8$ & 79.73 & 60.05 \\

    \midrule
    $\lambda_{KLD}=0.4$ & \textbf{80.04} & 58.04 &     $\lambda_{CE}=0.7$ & 77.53 & 60.78 \\

    \midrule
    $\lambda_{KLD}=0.5$ &77.83 & 60.99 &     $\lambda_{CE}=0.6$ & 77.49 & 57.61\\

    \midrule
    $\lambda_{KLD}=0.6$ & 78.09 & 58.36 &     $\lambda_{CE}=0.5$ & 76.72 & 58.74\\

    \midrule
    $\lambda_{KLD}=0.7$ &79.54 & 54.92 &     $\lambda_{CE}=0.4$ & 78.03 & 58.20\\

    \midrule

    $\lambda_{KLD}=0.8$ & 79.41 & 62.27 &     $\lambda_{CE}=0.3$ & 76.84 & 61.42\\

    \midrule
    $\lambda_{KLD}=0.9$ & 78.02 & 59.64 &     $\lambda_{CE}=0.2$ & 75.42 & 63.08\\

    \midrule
    $\lambda_{KLD}=1.0$ & 78.11 & 61.03  &     $\lambda_{CE}=0.1$ & 74.76 & 60.21\\
    
    \bottomrule

\end{tabular}
\caption{For this setting, the mIoU scores have been reported for Step 2 on the validation set of CrackSeg9k (source) and BuildCrack (Target). Our experiments show that $\lambda_{CE}=1.0$ and $\lambda_{KLD}=0.1$ gives the optimal performance on both the domains.}
  \label{tab:lambdaanaylsis}

\end{table}
\bibliographystyle{splncs04}
\bibliography{supp_ref}